# Robust Registration of Multimodal Remote Sensing Images Based on Structural Similarity

Yuanxin Ye, *Member, IEEE*, Jie Shan, *Senior Member, IEEE*, Lorenzo Bruzzone, *Fellow, IEEE*, and Li Shen

*Abstract*—Automatic registration of multimodal remote sensing data (e.g., optical, LiDAR, SAR) is a challenging task due to the significant non-linear radiometric differences between these data. To address this problem, this paper proposes a novel feature descriptor named the Histogram of Orientated Phase Congruency (HOPC), which is based on the structural properties of images. Furthermore, a similarity metric named HOPC$_{ncc}$ is defined, which uses the normalized correlation coefficient (NCC) of the HOPC descriptors for multimodal registration. In the definition of the proposed similarity metric, we first extend the phase congruency model to generate its orientation representation, and use the extended model to build HOPC$_{ncc}$. Then a fast template matching scheme for this metric is designed to detect the control points between images. The proposed HOPC$_{ncc}$ aims to capture the structural similarity between images, and has been tested with a variety of optical, LiDAR, SAR and map data. The results show that HOPC$_{ncc}$ is robust against complex non-linear radiometric differences and outperforms the state-of-the-art similarities metrics (i.e., NCC and mutual information) in matching performance. Moreover, a robust registration method is also proposed in this paper based on HOPC$_{ncc}$, which is evaluated using six pairs of multimodal remote sensing images. The experimental results demonstrate the effectiveness of the proposed method for multimodal image registration.

## I. INTRODUCTION

WITH the rapid development of geospatial information technology, remote sensing systems have entered an era where multimodal, multispectral, and multiresolution images can be acquired and jointly used. Due to the complementary information content of multimodal remote sensing images, it is necessary to integrate these images for Earth observation applications. Image registration, which is a fundamental preliminary task in remote sensing image processing, aligns two or more images captured at different times, by different sensors or from different viewpoints [1]. The accuracy of image registration has a significant impact on many remote sensing analysis tasks, such as image fusion, change detection, and image mosaic. Although remarkable progress has been made in automatic image registration techniques in the last few decades, their practical implementation for multimodal remote sensing image registration often requires manual selection of the control points (CPs) [2] (e.g., optical-to-Synthetic Aperture Radar (SAR) image or optical-to-Light Detection and Ranging (LiDAR) data registration) due to the significant geometric distortions and non-linear radiometric (intensity) differences between these images.

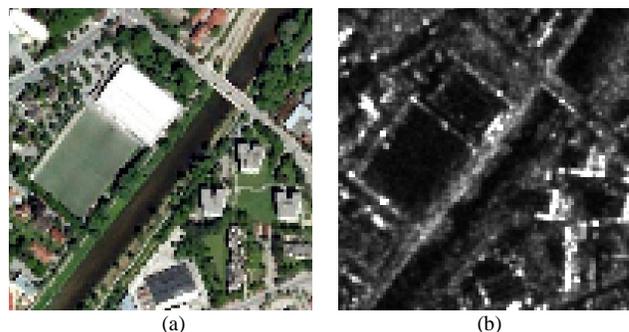

Fig. 1. Apparent non-linear radiometric differences between (a) optical image and (b) SAR image.

Current technologies enable the direct georeferencing of remote sensing images using physical sensor models and navigation devices aboard the platforms. These technologies can produce images that have an offset of only dozen or so pixels [3, 4] and are capable of removing nearly all the global geometric distortions from the images, such as obvious rotation and scale differences. The central difficulty for multimodal remote sensing image registration is related to non-linear radiometric differences. Fig. 1 shows a pair of optical and SAR images of the same scene with different intensity and texture patterns, which makes CP detection much more difficult than that under single-modal images. Therefore, the goal of this paper is to develop an effective registration method that is robust to non-linear radiometric differences between multimodal remote sensing images.

A typical automatic image registration process includes the

This paper is supported by the National key and Development Research Program of China (No. 2016YFB0501403 and No. 2016YFB0502603), the National Natural Science Foundation of China (No.41401369 and No.41401374), the Science and Technology Program of Sichuan, China (No. 2015SZ0046), and the Fundamental Research Funds for the Central Universities (No. 2682016CX083).

Y. Ye is with the Faculty of Geosciences and Environmental Engineering, Southwest Jiaotong University, Chengdu 610031, China, and also with the Department of Information Engineering and Computer Science, University of Trento, 38123 Trento, Italy (e-mail: yeyuanxin@home.swjtu.edu.cn).

L. Bruzzone is with the Department of Information Engineering and Computer Science, University of Trento, 38123 Trento, Italy (e-mail: lorenzo.bruzzone@ing.unitn.it).

J. Shan is with the School of Civil Engineering, Purdue University, West Lafayette, IN 47907 USA (e-mail: jshan@purdue.edu).

L. Shen is with the Faculty of Geosciences and Environmental Engineering, Southwest Jiaotong University, Chengdu 610031, China(e-mail: rsshenli@outlook.com).

following four steps: i) feature detection, ii) feature matching, iii) transformation model estimation, and iv) image resampling. Depending on the process adopted for registration, most multimodal remote sensing image registration methods can be classified into two categories: feature-based and area-based [1].

Feature-based methods first extract the remarkable features from both considered images, and then match them based on their similarities in order to achieve registration. Common image features include point features [5], line features [6], and region features [7]. Recently, local invariant features have been widely applied to image registration. Mikolajczyk et al. compared the performance of numerous local features for image matching and found that Scale Invariant Feature Transform (SIFT) [8] performed best for most of the tests [9]. Due to its invariance to image scale and rotation changes, SIFT has been widely used for remote sensing image registration [10-12]. However, SIFT is not effective for the registration of multimodal images, especially for optical and SAR images, because of its sensitivity to non-linear radiometric differences [13]. Past researchers proposed some new local invariant features based on SIFT, such as Speeded Up Robust Features [14], Oriented FAST and Rotated BRIEF [15], and Fast Retina Keypoint [16]. Although these new local features improve the computational efficiency, they are also vulnerable to complex radiometric changes. Fundamentally, the aforementioned feature-based methods mainly depend on detecting highly-repeatable common features between images, which can be difficult in multimodal images due to their non-linear radiometric differences [17]. Thus, these methods often do not achieve satisfactory performance for multimodal images.

Area-based methods (sometimes called template matching) usually use a template window of a predefined size to detect the CPs between two images. After the template window in an image is defined, the corresponding window over the other image is searched using certain similarity metrics. The centers of the matching windows are regarded as the CPs, which are then used to determine the alignment between the two images. Area-based methods have the following advantages compared with feature-based methods: (1) they avoid the step of feature detection, which usually has a low repeatability between multimodal images; and (2) they can detect CPs within a small search region because most remote sensing images are initially georeferenced up to an offset of several or dozens of pixels.

Similarity metrics play a decisive role in area-based methods. Common similarity metrics include the sum of squared differences (SSD), the normalized cross correlation (NCC), and the mutual information (MI). SSD is probably the simplest similarity metric because it detects CPs by directly computing the intensity differences between two images. However, SSD is quite sensitive to radiometric changes despite its high computational efficiency. NCC is a very popular similarity metric and is widely applied to the registration of remote sensing images because of its invariance to linear intensity variations [18, 19]. However, NCC is vulnerable to non-linear radiometric differences [20]. In contrast, MI is more robust to complex radiometric changes and is extensively used in multimodal image registration [21-23]. Unfortunately, MI is computationally expensive because it must compute the joint histogram of each window to be matched [19] and is very sensitive to the window size for template matching [20]. These drawbacks limit its broad use in multimodal remote sensing image registration. In general, all three similarity metrics cannot effectively handle significant radiometric distortions between images because they are mainly applied on image intensities. Past researchers improved the performance of registration by applying these metrics to image descriptors such as gradient features [24] and wavelet-like features [25, 26]. However, these features are difficult to use for reflecting the common properties of multimodal images.

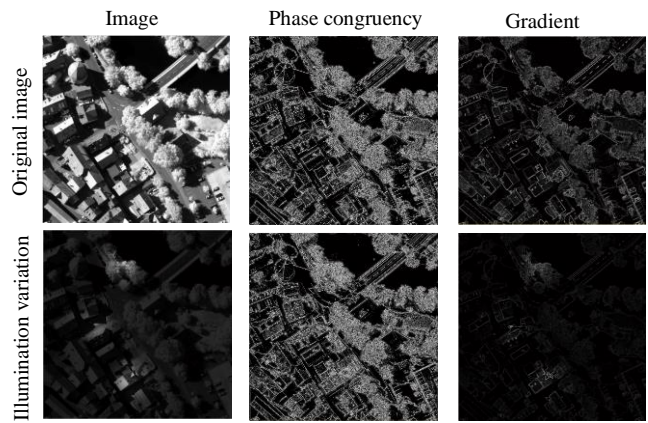

Fig. 2 Comparison of phase congruency with gradient.

Recently, in multimodal medical image processing, structure and shape features have been integrated as similarity metrics for image registration and have achieved better performance than traditional similarity metrics [27-30]. These methods are based on the assumption that structure and shape properties are preserved across different modalities and are relatively independent of radiometric changes. Inspired by that assumption, the work presented in this paper explores the performance of structural properties for multimodal remote sensing image registration. As shown in Fig. 1, the contour structures and geometric shapes are quite similar between the optical and SAR images despite their very different intensity characteristics. To address this issue, a novel similarity metric is proposed in this paper to exploit the similarity between structural features to deal with the non-linear radiometric differences between multimodal images. In general, structural features can be represented by gradient information of images, but gradient information is sensitive to the radiometric changes between images. In contrast, the phase congruency feature has been demonstrated to be more robust to illumination and contrast variation [31] (see Fig. 2). This characteristic makes it insensitive to radiometric changes. However, the conventional phase congruency model can only obtain its amplitude, which is insufficient for structural feature description [32]. This paper therefore extends the conventional phase congruency model to build its orientation representation. Both amplitude and orientation then are used to construct a novel feature descriptor that can capture the structures of images. This descriptor, named the Histogram of Orientated Phase Congruency (HOPC), can be efficiently calculated in a dense manner over the whole image. The idea of HOPC is inspired from the histogram of oriented gradient (HOG), which has been very successful in target recognition [33]. The HOPC descriptor reflects the structural properties of images, which are relatively independent of the particular intensity distribution pattern across two images. The HOPC descriptor can be extracted for each image separately and then directly compared across images using a simple intensity metric such as NCC. Therefore, the NCC of the HOPC descriptors is used as the similarity metric (named HOPC$_{ncc}$), and a fast template matching scheme is designed to detect CPs between images. In addition, in accordance with the characteristics of remote sensing images, an automatic registration method is designed based on HOPC$_{ncc}$. The main contributions of this paper are as follows.

(1) Extension of the phase congruency model to build its orientation representation.

(2) Development of a novel similarity metric (named HOPC$_{ncc}$) based on both the amplitude and orientation of phase congruency to address the non-linear radiometric differences between multimodal images as well as a fast template matching scheme to detect CPs between images and an automatic registration method for multimodal remote sensing images based on HOPC$_{ncc}$.

This paper extends a preliminary version of this work [32] by adding (1) a detailed principled derivation of HOPC$_{ncc}$; (2) a detailed analysis of the effects of the various parameters on HOPC$_{ncc}$; (3) an effective multimodal registration method based on HOPC$_{ncc}$; and (4) a more thorough evaluation process through the use of a larger quantity of multimodal remote sensing data. The code of the proposed method can be downloaded in this website[1].

The remainder of this paper is organized as follows. Section II describes the proposed similarity metric (HOPC$_{ncc}$) for multimodal registration. Section III proposes a robust registration method based on HOPC$_{ncc}$. Section IV analyzes the parameter sensitivity of HOPC$_{ncc}$ and compares it with the state-of-the-art similarity metrics by using various multimodal remote sensing datasets. Section V evaluates the proposed registration method based on HOPC$_{ncc}$. Section VI presents the conclusions and recommendations for future work.

## II. HOPC$_{NCC}$: STRUCTURAL SIMILARITY METRIC

Given a master image $I_1(x, y)$ and a slave image $I_2(x, y)$, the aim of image registration is to find the optimal geometric transformation model that maximizes the similarity metric between $I_1(x, y)$ and the transformed $I_2(x, y)$ named $I_2(T(x, y))$, which can be expressed as:

$$\hat{T}(x, y) = \underset{T(x, y)}{\arg\max}\left[\Psi(I_2(T(x, y)), I_1(x, y))\right] \quad (1)$$

where $T(x, y)$ is the geometric transformation model, and

$\Psi(.)$ is the similarity metric.

In this section, we present a novel structural descriptor named HOPC and define the similarity between two images on the basis of HOPC. The proposed descriptor is based on the assumption that multimodal images share similar structural properties despite having different intensity and texture information. First, the phase congruency model is extended to generate its orientation representation, which then is used to construct the structural similarity metric HOPC$_{ncc}$; and a fast template matching scheme for this metric is designed to detect the CPs between images.

### A. Analysis of Importance of Phase

Many feature detectors and descriptors are based on gradient information, such as Sobel, Canny [34], and SIFT. As already mentioned, these operators are usually sensitive to image illumination and contrast changes. By comparison, the phase information of images is more robust to these changes. Let us consider an image $I(x)$, and its Fourier transform $F(\omega) = |F(\omega)|e^{-j\theta(\omega)}$. $|F(\omega)|$ and $\theta(\omega)$ are the amplitude and the phase of the Fourier transform, respectively. Oppenheim et al. [35] analyzed the phase function for image processing and found that the phase of an image is more important than the amplitude. This conclusion is clearly illustrated in Fig. 3. Images $a$ and $b$ are first analyzed with the Fourier transform to obtain the phase $\theta(a)$ and amplitude $|F(a)|$ of image $a$, as well as the phase $\theta(b)$ and amplitude $|F(b)|$ of image $b$, respectively. Then, $\theta(a)$ and $|F(b)|$ are used to synthetize a new image $a_b$ by applying inverse Fourier transform. $\theta(b)$ and $|F(a)|$ also are composed as a new image $b_a$ through the same procedure. It can be clearly observed that $a_b$ and $b_a$ both mainly present the information of the image that provides the phase, which shows that the contour and structural features of the images are mainly provided by the phase.

### B. Phase Congruency

Since phase has been demonstrated to be important for image perception, it is natural to use it for feature detection. Phase congruency is a feature detector based on the local phase of an image, which postulates that features such as corners and edges are present where the Fourier components are maximally in phase. Morrone and Burr [36] have demonstrated that this model is conformed to the human visual perception of image features. Phase congruency is invariant to illumination and contrast changes because of its independence of the amplitude of signals [31]. Given a signal $f(x)$, its Fourier series expansion is $f(x) = \sum_n A_n \cos(\phi_n(x))$, where $A_n$ is the amplitude of the $n^{th}$ Fourier component, and $\phi_n$ is the local phase of the Fourier component at position $x$. The phase congruency of this signal is defined as

$$PC_1(x) = \max_{\varphi(x) \in [0, 2\pi)} \left\{ \frac{\sum_n A_n(x) \cos(\phi_n(x) - \bar{\phi}(x))}{\sum_n A_n(x)} \right\} \quad (2)$$

where $\bar{\phi}(x)$ is the amplitude weighted mean local phase of all the Fourier terms at position $x$ to maximize this equation. Since this model cannot accurately locate features in noisy and blurred images, Kovesi have improved the calculation model of phase congruency by using log Gabor wavelets over multiple scales and orientations [31]. In the frequency domain, the log Gabor function is defined as

$$g(\omega) = \exp\left(\frac{-(\log(\omega / \omega_0))^2}{2(\log(\sigma_\omega / \omega_0))}\right) \quad (3)$$

where $\omega_0$ is the central frequency and $\sigma_\omega$ is the related width parameter. The corresponding filter of the log Gabor wavelet in the spatial domain can be achieved by applying the inverse Fourier transform. The "real" and "imaginary" components of the filter are respectively referred as the log Gabor even-symmetric $M_{no}^e$ and odd-symmetric $M_{no}^o$ wavelets (see Fig. 4). Given an input image $I(x, y)$, its convolution results with the two wavelets can be regarded as a response vector.

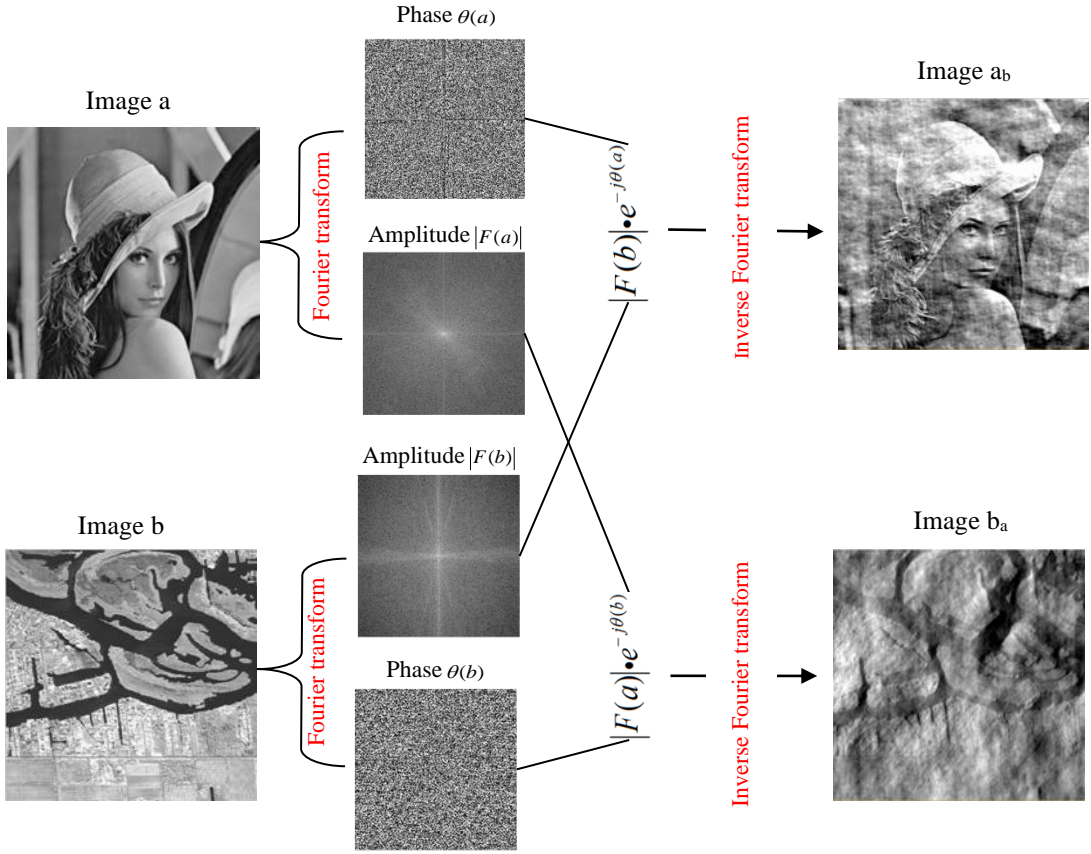

Fig. 3 Illustrative example of importance of image phase.

$$\left[ e_{no}(x, y), o_{no}(x, y) \right] = \left[ I(x, y) * M_{no}^e, I(x, y) * O_{no}^e \right] \quad (4)$$

where $e_{no}(x, y)$ and $o_{no}(x, y)$ are the responses of $M_{no}^e$ and $M_{no}^o$ at scale $n$ and orientation $o$. The amplitude $A_{no}$ and phase $\phi_{no}$ of the transform at a wavelet scale $n$ and orientation $o$ are given by

$$A_{no} = \sqrt{e_{no}(x, y)^2 + o_{no}(x, y)^2}$$
$$\phi_{no} = atan2(e_{no}(x, y), o_{no}(x, y)) \quad (5)$$

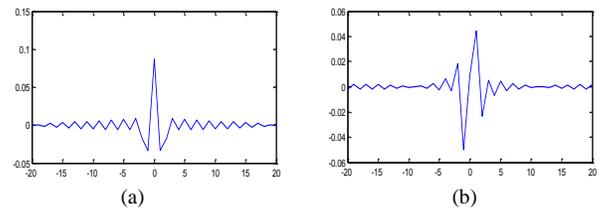

Fig. 4 Log Gabor wavelets. (a) Even-symmetric wavelet. (b) Odd-symmetric wavelet.

Considering the noise and blur of images, the improved phase congruency model (named PC$_2$) proposed by Kovesi is defined as

$$PC_2(x,y) = \frac{\sum_n \sum_o W_o(x,y) \lfloor A_{no}(x,y)\Delta\Phi_{no}(x,y) - T \rfloor}{\sum_n \sum_o A_{no}(x,y) + \varepsilon} \quad (6)$$

where $(x,y)$ indicates the coordinates of the point in an image, $W_o(x,y)$ is the weighting factor for the given frequency spread, $A_{no}(x,y)$ is the amplitude at $(x,y)$ for the wavelet scale $n$ and orientation $o$, $T$ is a noise threshold, $\varepsilon$ is a small constant to avoid division by zero, $\lfloor \ \rfloor$ denotes that the enclosed quantity is equal to itself when its value is positive or zero otherwise. $\Delta\Phi_{no}(x,y)$ is a more sensitive phase deviation defined as

$$A_{no}(x,y)\Delta\Phi_{no}(x,y) = \left( e_{no}(x,y).\overline{\phi}_e(x,y) + o_{no}(x,y).\overline{\phi}_o(x,y) \right)$$
$$- \left| e_{no}(x,y).\overline{\phi}_o(x,y) - o_{no}(x,y).\overline{\phi}_e(x,y) \right| \quad (7)$$

where $\overline{\phi}_e(x,y) = \sum_n \sum_o e_{no}(x,y) / E(x,y)$ and $\overline{\phi}_o(x,y) = \sum_n \sum_o o_{no}(x,y) / E(x,y)$. The term $E(x,y)$ is the local energy function and is expressed as $E(x,y) = \sqrt{\left( \sum_n \sum_o e_{no}(x,y) \right)^2 + \left( \sum_n \sum_o o_{no}(x,y) \right)^2}$.

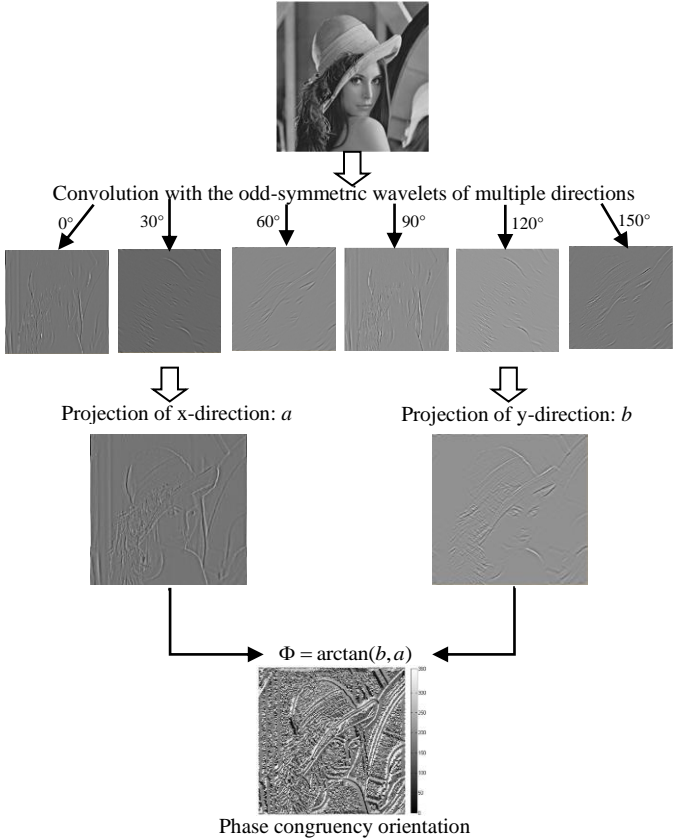

Fig. 5 Illustrative example of calculation of the orientation of phase congruency.

## C. Orientation of Phase Congruency

The above conventional phase congruency model only considers feature amplitudes of pixels (like gradient amplitudes). However, it cannot achieve their feature orientations (like gradient orientations) that represent the significant directions of feature variation. This traditional phase congruency model cannot effectively describe the feature distribution of the local regions of images. Thus, it is insufficient to use only the amplitude of phase congruency to construct robust feature descriptors. Taking the SIFT operator as an example, apart from gradient amplitudes, gradient orientations are also used to build the feature descriptors. Therefore, we extend the phase congruency model to build its orientation representation for constructing the feature descriptor.

As mentioned in the above subsection, the phase congruency feature is computed by the log Gabor odd-symmetric and even-symmetric wavelets. The log Gabor odd-symmetric wavelet is a smooth derivative filter (see Fig. 4(b)) which can compute the image derivative in a single direction (like gradients) [37]. Considering that the log Gabor odd-symmetric wavelets of multiple directions are used in the computation of phase congruency, the convolution result of each directional wavelet can be projected onto the horizontal direction (x-direction) and vertical direction (y-direction), yielding the x-direction derivative $a$ and the y-direction derivative $b$ of the images, respectively (see Fig. 5). The orientation of phase congruency is defined as

$$a = \sum_\theta (o_{no}(\theta)\cos(\theta))$$
$$b = \sum_\theta (o_{no}(\theta)\sin(\theta)) \quad (8)$$
$$\Phi = \arctan(b,a)$$

where $\Phi$ is the orientation of phase congruency and $o_{no}(\theta)$ denotes the convolution results of log Gabor odd-symmetric wavelet at orientation $\theta$. Fig. 5 illustrates the process of calculating the orientation of phase congruency, which has a domain in the range [$0^o$, $360^o$).

## D. Structural Feature Descriptor

The aim of the work in this paper is to find a descriptor that is as independent as possible of the intensity patterns of images from different modality. In this subsection, a feature descriptor named HOPC is proposed which uses both the amplitude and orientation of phase congruency. The HOPC descriptor captures the structural properties of images. It is inspired from HOG, which can effectively describe local object appearances and shapes through the distribution of the gradient amplitudes and orientations of local image regions. HOG has been successfully applied to object recognition [38], image classification [39] and image retrieval [40] because it represents the shape and structural features of images. This descriptor characterizes the structural properties of images using gradient information. Phase congruency, similarly to gradients, also reflects the significance of the features of local image regions. Moreover, this model is more robust to image illumination and contrast changes. As such, the amplitude and orientation of phase congruency are utilized to build the HOPC descriptor based on the framework of HOG.

As shown in Fig. 6, HOPC is calculated based on the evaluation of a dense grid of well-normalized local histograms of phase congruency orientations over a template window selected in an image. The main steps for extracting the HOPC descriptor are described below.

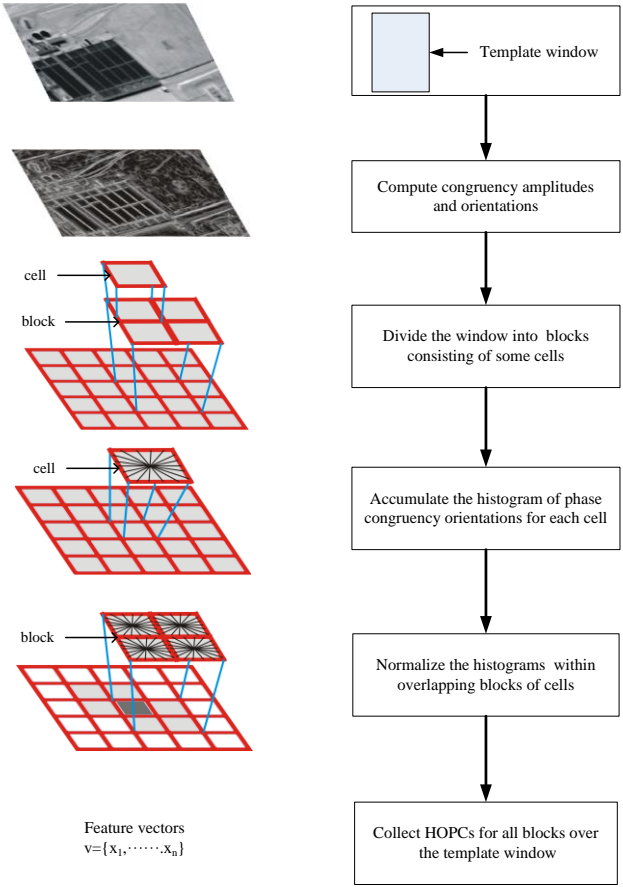

Feature vectors
v={x₁,……,xₙ}

Fig. 6 Main processing chain for calculating the HOPC descriptor.

(1) The first step selects a template window with a certain size in an image, and then computes the phase congruency amplitude and orientation for each pixel in this template window, which provides the feature information for HOPC.

(2) The second step divides the template window into overlapping blocks, where each block consists of m × m spatial regions, called "cells," each containing n × n pixels. This process defines the fundamental framework of HOPC.

(3) The third step accumulates a local histogram of the phase congruency orientations of all the pixels within the cells of each block. Each cell is first divided into a number of orientation bins, which are used to form the orientation histograms. A Gaussian spatial window is applied to each pixel before accumulating orientation votes into the cells in order to emphasize the contributions of the pixels near the center of the cell. Then, the histograms are weighted by phase congruency amplitudes using a trilinear interpolation method. The histograms for the cells in each block are normalized by the L2 norm to achieve a better robustness to illumination changes. This process produces the HOPC descriptor for each block. It should be noted that the phase congruency orientations need to be limited to the range [ 0º , 180º ) to construct the orientation histograms, which can handle the intensity inversion between multimodal images.

(4) The final step collects the HOPC descriptors of all the blocks in the template window into a combined feature vector, which can be used for template matching.

The HOPC descriptor can capture the structural features of images and is more robust to illumination changes compared with the HOG descriptor. Fig. 7 shows an example of the HOPC descriptors computed from a local region of two images with significant illumination variations. There are more similarities between the HOPC descriptors than the HOG descriptors.

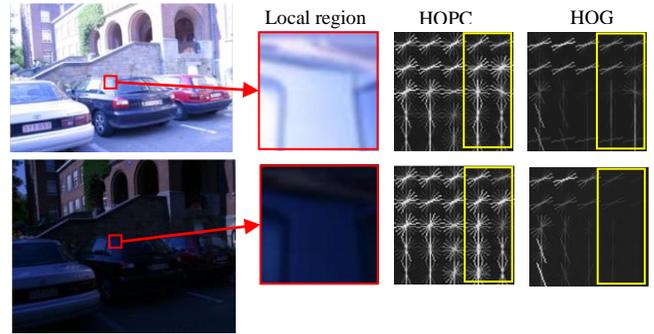

Fig. 7 Comparison of HOPC with HOG. It is possible to see that the HOPC descriptors are more robust to illumination changes than the HOG descriptors.

### E. Similarity Metric Based on Structural Properties

As mentioned above, HOPC is a feature descriptor that captures the internal structures of images. Since structural properties are relatively independent of intensity distribution patterns of images, this descriptor can be used to match two images having significant non-linear radiometric differences as long as they both have similar shapes. Therefore, the NCC of the HOPC descriptors is taken as the similarity metric (named HOPC_ncc) for image registration, which is defined as

$$HOPC_{ncc} = \frac{\sum_{k=1}^{n}(V_A(k)-\bar{V}_A)(V_B(k)-\bar{V}_B)}{\sqrt{\sum_{k=1}^{n}(V_A(k)-\bar{V}_A)^2\sum_{k=1}^{n}(V_B(k)-\bar{V}_B)^2}} \quad (9)$$

where $V_A$ and $V_B$ are HOPC descriptors of the image region $A$ and image region $B$ respectively. $\bar{V}_A$ and $\bar{V}_B$ denote the means of $V_A$ and $V_B$ , respectively.

### F. Fast Matching Scheme

During the template matching processing, a template window moves pixel-by-pixel within a search region or an image. For each pair of template windows to be matched, we have to compute its $HOPC_{ncc}$. Since most of the pixels overlap between adjacent template windows, this requires many repetitive computations. To address this issue, a fast matching scheme is designed for $HOPC_{ncc}$.

The CP detection using $HOPC_{ncc}$ includes two steps: extracting the HOPC descriptors and computing the NCC between such descriptors. The first step spends the most time in the matching process. To extract the HOPC descriptor, the template window is divided into some overlapping blocks, and the descriptors for each of these blocks are collected to form the final dense descriptor. Therefore, a block can be regarded as the fundamental element of the HOPC descriptor. In order to reduce the computational time of template matching, we define a block region centered on each pixel in an image (or a search region), and extract the HOPC descriptor of each block (hereafter referred to as the block-HOPC descriptor). Each pixel will then have a block-HOPC descriptor that forms the 3D descriptors for the whole image, which is called the block-HOPC image. Then the block-HOPC descriptors are collected at an interval of several pixels (such as a half block width[2]) to generate the HOPC descriptor for the template window. Fig. 8 illustrates the fast computing scheme.

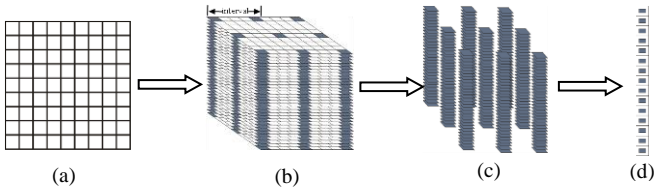

Fig. 8 Illustration of the fast computing scheme for the HOPC descriptor. (a) Image. (b) Block-HOPC image. (c) Block-HOPC descriptors at a certain interval. (d) Final HOPC descriptor.

This scheme can eliminate much of the repetitive computation between adjacent template windows. Let us now compare the computational efficiency of our matching scheme with the traditional matching scheme. For a template window ($N \times N$ pixels) that has a moving search region[3] with a size of $M \times M$ pixels, the traditional scheme takes $O\left(M^2 N^2\right)$ operations because the template window slides pixel-by-pixel across the search region. Differently from the traditional scheme, the computational time taken from our scheme mainly includes the two parts: (1) extraction of the block-HOPC descriptors for all pixels in the whole search region ($(M+N)^2$ pixels); (2) collection of the block-HOPC descriptors at intervals of a half block width for all of the template windows used to match. The computational cost of the latter can almost be ignored compared to that of the former because it simply assembles the block-HOPC descriptor at a certain interval sampling. The former needs $O\left((M+N)^2\right)$ operations for extracting the block-HOPC descriptor for each pixel in the whole search region. Compared with the traditional scheme, our scheme has a significant computational advantage in the large size of template window or search region. Fig. 8 shows the run times for the two schemes versus the size of template window and search region, when 200 interest points are matched. One can observe that our scheme requires much less time than the traditional scheme, and this advantage becomes more and more obvious by increasing the template window and search region size.

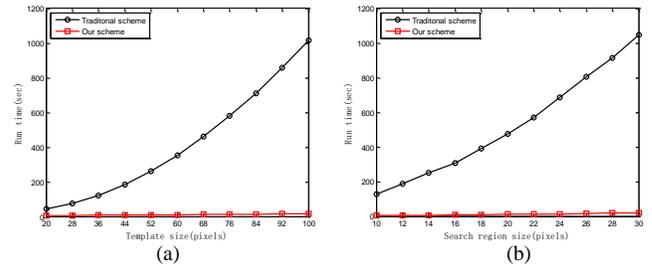

Fig. 8 Comparison of run time taken from the traditional matching scheme and our scheme using $HOPC_{ncc}$. (a) Run time versus the template size, where the size of the search region is 20×20 pixels. (b) Run time versus the search region size, where the template size is 68×68 pixels.

### III. MULTIMODAL REGISTRATION METHOD BASED ON $HOPC_{NCC}$

In this section, a novel robust image registration method is introduced for multimodal images based on $HOPC_{ncc}$, which consists of the following six steps. Fig. 10 shows the flowchart of the proposed method.

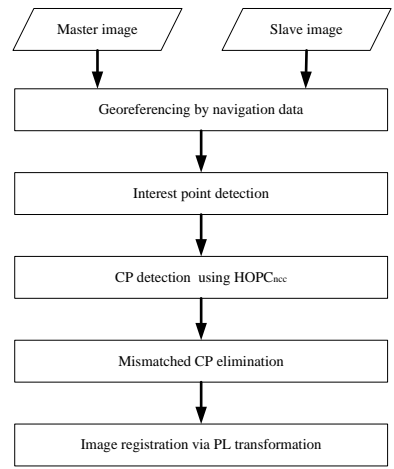

Fig. 10 Flowchart of the proposed image registration method.

(1) The master and slave images are first coarsely rectified using the direct georeferencing techniques to remove their obvious translation and rotation differences. Then, the two images are resampled to the same ground sample distance

---

[2] This makes the adjacent blocks have the overlap of 50% to build the HOPC descriptor.

[3] This refers to the moving range of center pixel of template window.

(GSD) to eliminate possible resolution differences.

(2) In order to evenly distribute the CPs over the image, the block-based Harris operator [13] is used to detect the interest points in the master image. The image is first divided into $n \times n$ non-overlapping blocks, and the Harris values are computed for each block. Then, the Harris values are ranked from the largest to the smallest in each block, and the top $k$ points are selected as the interest points.

(3) Once a set of interest points is extracted in the master image, HOPC$_{ncc}$ is used to detect the CPs using a template matching scheme in a small search window of the slave image, which is determined through the georeferencing information of the images. To increase the robustness of the image matching, a bidirectional matching technique [41] is applied, which includes two steps (forward matching and backward matching). In the forward step, for an interest point $p_1$ in the master image, its match point $p_2$ is found by the maximum of HOPC$_{ncc}$ between the template window in the master image and the search window in the slave image. In the backward step, the match point of $p_2$ is found in the master image by the same method. Only when the two matching steps achieve consistent results, the matched point pair $(p_1, p_2)$ is considered as CPs.

(4) Due to existing uncertainty factors, such as occlusion and shadow, the obtained CPs are not error-free. Large CP errors are eliminated using a global consistency check method based on a global transformation [5]. The transformation model chosen is vital for the consistency check and depends on the types of relative geometric deformations between images. In this paper, the projective transformation model is chosen for the consistency check because it can effectively handle common global transformation (translation, rotation, scale, and shear) [42].

(5) Mismatched CPs are removed by an iterative refining procedure. A projective transformation model is first set up using the least squares method with all the CPs. The residuals and the root mean-square error (RMSE) of CPs then are computed, and the CP with the largest residual is removed. The above process is repeated until the RMSE is less than a given threshold (e.g., 1 pixel).

(6) After the CPs with large errors are removed, it is necessary to determine a transformation model to rectify the slave image. A piecewise linear (PL) transformation model [43] is chosen to address the local distortions caused by terrain relief. This model first divides the images into triangular regions using Delaunay's triangulation method [44], and an affine transformation (see (10)) is applied to map each triangular region in the slave image onto the corresponding region in the master image [45].

$$x_1 = a_0 + a_1 x_2 + a_2 y_2$$
$$y_1 = b_0 + b_1 x_2 + b_2 y_2$$
(10)

where $(x_1, y_1)$ and $(x_2, y_2)$ are the coordinates of the CPs

in the master and slave images, respectively.

## IV. Experimental Results: HOPC$_{NCC}$ Matching Performance

In this section, the matching performance of HOPC$_{ncc}$ is evaluated using different types of multimodal remote sensing images by considering three metrics: the similarity curve, the correct match ratio (CMR), and the computational efficiency. The experiments mainly have two objectives: (1) test the influences of the various parameters for HOPC$_{ncc}$ and (2) compare HOPC$_{ncc}$ with the state-of-the-art similarity metrics such as NCC and MI. In the experiments conducted, MI is computed by a histogram with 32 bins because it achieves the optimal matching performance for the datasets used. In addition, since HOPC$_{ncc}$ uses the framework of HOG to build the similarity metric, the HOG descriptor is also integrated as a similarity metric for the comparison with HOPC$_{ncc}$. Based on our analysis of the literature, to the authors' knowledge, the HOG descriptor has not been previously used as a similarity metric for multimodal remote sensing image registration in a template matching scheme. The NCC of the HOG descriptors is used as the similarity metric (named HOG$_{ncc}$) for image matching. It is empirically found that the original parameter setting [33] of the HOG descriptor could not be efficiently applied to multimodal remote sensing image matching, which is likely because these parameters are designed for target detection only. Therefore, HOG$_{ncc}$ is set to the same parameters as HOPC$_{ncc}$ (see Section IV-C) for image matching in order to make a fair comparison. The test data, implementation details, and experimental analysis are as follows.

### A. Description of Datasets

Two categories of multimodal remote sensing image pairs (synthetic images with non-linear radiometric differences and real multimodal images) are used to evaluate the effectiveness of HOPC$_{ncc}$.

*Synthetic Datasets*

Two different types of intensity warped models are used to generate the synthetic images. A high resolution image (1382 $\times$ 1382 pixels) located in an urban area is used to perform the synthetic experiment. The master and slave images are simulated by applying a spatially-varying intensity warped model (see (11)) and a piecewise linear mapping function to the image, respectively. Moreover, a Gaussian noise with mean $\mu$=0 and variance $\sigma^2$=0.2 is imposed on the slave image.

$$I(x, y) = I(x, y) . (0.1 + \frac{1}{K} \sum_{k=1}^{K} e^{-\|x,y\| - \mu_k\|^2 / (2(180)^2)})$$
(11)

where, $I(x, y)$ denotes the image rescaled to [0,1]. The last term in the brackets models the locally-varying intensity field with a mixture of $K$ randomly centered Gaussians [46] with $K$ set to 3 to generate the synthetic image.

The spatially-varying intensity warped model generates an image having non-uniform illumination and contrast changes, while the piecewise linear mapping function introduces a non-linear radiometric distortion model to warp the image.

Such radiometric distortion models have been applied for the simulation of multimodal matching in the literature [19, 20, 46]. Fig. 11 shows the process used for generating the synthetic master and slave images which present the significant radiometric differences.

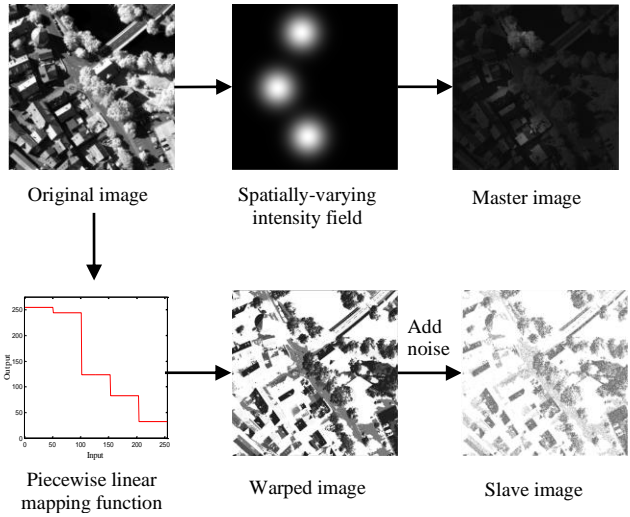

Fig. 11 Illustration of the process used to generate the synthetic images used in the experiments.

*Real Datasets*

Ten sets of real multimodal image pairs are used to evaluate the effectiveness of HOPC_ncc. These images are divided into four categories: Visible-to-Infrared (Visib-Infra), LiDAR-to-Visible (LiDAR-Visib), Visible-to-SAR (Visib-SAR), and Image-to-Map (Img-Map). The tested image pairs are a variety of medium resolution (30m) and high-resolution (0.5m to 3m) images that included different terrains and both urban and suburban areas. All of the image pairs have been systematically corrected by using their physical models, and each image pair is respectively resampled into the same GSD. Consequently, there are only a few obvious translation, rotation, and scale differences between the master and slave images. However, significant radiometric differences are expected between images because they are acquired by different imaging modalities and at various spectra. Fig 19 shows the test data, and Table I presents descriptions of the data. The characteristics of each test set are as follows.

*Visible-to-Infrared:* Visib-Infra 1 and Visib-Infra 2 are visible and infrared data which include a pair of high-resolution images and a pair of medium-resolution images. The high-resolution images represent an urban area, while the medium-resolution images cover a suburban area.

*LiDAR-to-Visible:* Three pairs of LiDAR and visible data are selected for the experiments. LiDAR-Visib 1 and LiDAR-Visib 2 are two pairs of interpolated raster LiDAR intensity and visible images covering urban areas with high buildings. They have obvious local geometric distortions caused by the relief displacement of buildings. Moreover, the LiDAR intensity images have significant noise, which increase the difficulty of matching. LiDAR-Visib 3 includes a pair of interpolated raster LiDAR height and visible images. Large differences can be observed from the intensity characteristics of the two images (see Fig. 19), which make matching the two images quite challenging.

*Visible-to-SAR:* Visib-SAR 1 to Visib-SAR 3 are composed of visible and SAR images. Visib-SAR 1 contains a pair of medium-resolution images located in a suburban area. Visib-SAR 2 and Visib-SAR 3 are high resolution images covering urban areas with high buildings, thus resulting in obvious local distortions. Additionally, there is a temporal difference of 14 months between the images in Visib-SAR 3, and some ground objects therefore changed during this period. These differences make it very difficult to match the two images.

*Image-to-Map:* Img-Map 1 and Img-Map 2 are two pairs of visible images and map data downloaded from Google Maps. The map data have been rasterized. Since both pairs of data represent urban areas with high buildings, local distortions are evident between the two images of each pair. In addition, the radiometric properties between the visible images and the map data are almost completely different. As shown in Fig. 19, the texture information of the maps is much poorer than that of the images, and there are also some labeled texts in the map. Therefore, it is very challenging to detect the CPs between the two data.

### B. Implementation Details and Evaluation Criteria

First, the block-based Harris operator (see Section III) is used to detect the interest points in the master image, where the image is divided into $10 \times 10$ non-overlapping blocks, and two interest points are extracted from each block, for a total of 200 interest points. Then NCC, MI, HOG_ncc, and HOPC_ncc are applied to detect the CPs within a search region of a fixed size ($20 \times 20$ pixels) of the slave image using a template matching strategy, after which the similarity surface is fitted using a quadratic polynomial to determine the subpixel position [10].

CMR is chosen as the evaluation criterion and is calculated as CMR=CM/C , where CM is the number of correctly matched point pairs in the matching results, and C is the total number of match point pairs. The matched point pairs with localization errors smaller than a given threshold value are regarded as the CM. For the synthetic datasets, a small threshold value (0.5 pixels) is used to determine the CM because of the known exact geometric distortions between the master and slave images. For the real datasets, we determine the CM by selecting a number of evenly distributed check points for each image pair. In general, the check points are determined by manual selection. However, for some multimodal image pairs, especially for LiDAR-Visib and Visib-SAR, it is very difficult to locate the CPs precisely by visual inspection due to their varying intensity and texture characteristics. Accordingly, different strategies are designed to select the check points based on the characteristics of the datasets. For Visib-Infra, the images have relatively more similar radiometric characteristics than those of other datasets, and a set of 40-60 evenly distributed check points are manually selected between the master and slave images. For the other datasets, especially for the LiDAR and SAR data, HOPC$_{ncc}$ has been used to detect 200 evenly distributed CPs between images by a large template size (200×200 pixels) because the experiments show that a larger template window can achieve higher CMR values (see Section IV-E). Then, the CPs with large errors are eliminated using the global consistency check method described in Section III. Finally 40-60 CPs with the least residuals are selected as the check points.

TABLE I
DESCRIPTION OF DATASETS USED IN THE MATCHING EXPERIMENTS

| Category | No. | Image pair | Size and GSD | Date | Image Characteristic |
|---|---|---|---|---|---|
| Visib-Infra | 1 | Daedalus visible<br>Daedalus infrared | 512×512, 0.5m<br>512×512, 0.5m | 04/2000<br>04/2000 | Urban area |
| | 2 | Landsat 5 TM band 1(visible)<br>Landsat 5 TM band 4(infrared) | 800×800, 30m<br>800×800, 30m | 09/2001<br>3/2002 | Suburban area, temporal difference of 6 months |
| LiDAR-Visib | 1 | LiDAR intensity<br>WorldView2 visible | 600×600, 2m<br>600×600, 2m | 10/2010<br>10/2011 | Urban area with high buildings, significant local distortions, temporal difference of 12 months, and significant noise in the LiDAR data |
| | 2 | LiDAR intensity<br>WorldView2 visible | 621×617, 2m<br>621×621, 2m | 10/2010<br>10/2011 | Urban area, temporal difference of 12 months, and serious noise in the LiDAR data |
| | 3 | LiDAR height<br>Airborne visible | 524×524, 2.5m<br>524×524, 2.5m | 06/2012<br>06/2012 | Urban area, and large difference in intensity characteristics |
| Visib-SAR | 1 | Landsat 5 TM band 3<br>TerraSAR-X | 600×600, 30m<br>600×600, 30m | 05/2007<br>03/2008 | Suburban area , and significant noise in the SAR data |
| | 2 | Image from Google Earth<br>TerraSAR-X | 528×524, 3m<br>534×524, 3m | 11/2007<br>12/2007 | Urban area, and significant noise in the SAR data |
| | 3 | Image from Google Earth<br>TerraSAR-X | 628×618, 3m<br>628×618, 3m | 03/2009<br>01/2008 | Urban area, local distortions, temporal difference of 14 months, and significant noise in the SAR data. |
| Img-Map | 1 | Image from Google Maps<br>Map from Google Maps | 700×700, 0.5m<br>700×700, 0.5m | N/A | Urban area with high buildings, obvious local distortions, large differences in intensity characteristics |
| | 2 | Image from Google Maps<br>Map from Google Maps | 621×614, 1.5m<br>621×614, 1.5m | N/A | Urban area with high buildings, obvious local distortions, large differences in intensity characteristics |

Once the check points are selected, the projective transformation model computed using these points is employed to calculate the localization error of each matched point pair. The threshold value of the error is set to 1.0 pixel to determine the CM for the image pairs of Visib-Infra 2 and Visib-SAR 1 because they have few local distortions. For the other high resolution image pairs, the threshold value is set to 1.5 pixels for achieving higher flexibility since their rigorous geometric transformation relationships are usually unknown and a projective transformation model can only pre-fit the geometric distortions.

### C. Parameter Tuning

This subsection systematically analyzes the effects of various parameters on the performance of HOPC$_{ncc}$. HOPC$_{ncc}$ is constructed using blocks having a $\alpha$ degree of overlap. Each block consists of $m \times m$ cells containing $n \times n$ pixels, and each cell is divided into $\beta$ orientation bins. Thus, $\alpha$, $m$, $n$, $\beta$ are the parameters to be tuned; and their influences are tested on the ten sets of multimodal images described in Table I. In this experiment, HOPC$_{ncc}$ is used to detect the CPs between the images by a template matching scheme, where the template size is set to 100 × 100 pixels. The average CMR is used to assess the influence of the various parameters because multiple sets of data are used in the experiment.

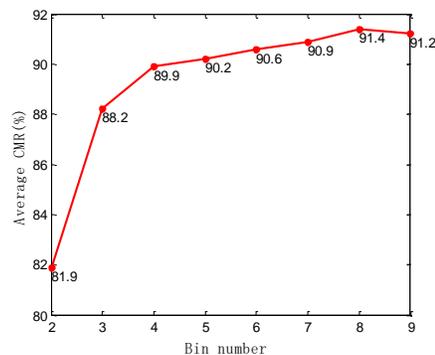

Fig. 12 Average CMR values versus the orientation bin number $\beta$ .

We first test the influence of the number of orientation bins $\beta$ on HOPC$_{ncc}$, when HOPC$_{ncc}$ is constructed by 3×3 cell blocks of 4×4 pixel cells, and the overlap $\alpha$ between blocks is set to a half-block width ($\alpha$=1/2). Fig. 12 shows the average CMR values versus the number of orientation bins. It can be observed that the average CMR value generally increases with the number of orientation bins. It reaches the maximum value when the bin number is 8. Therefore $\beta$=8 is regarded as a good-selected for HOPC$_{ncc}$.

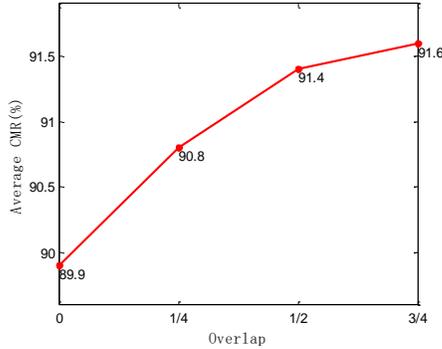

Fig. 13 Average CMR values versus the degree of overlap $\alpha$.

In the procedure for building HOPC$_{ncc}$, the blocks are overlapped so that each cell in a block contributes several components to the final descriptor. Therefore, the degree of overlap affects the performance of HOPC$_{ncc}$. Fig. 14 shows that the average CMR value increases as the amount of overlap in the range between 0 and 3/4 block widths is increased, but the differences between the overlaps of 1/2 and 3/4 block widths is small. Since a larger overlap is more time-consuming, a half block width ($\alpha$=1/2) is chosen as the default setting for HOPC$_{ncc}$.

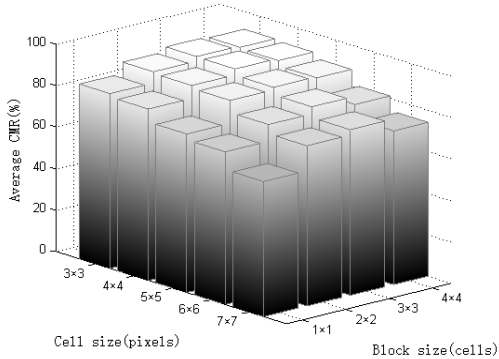

Fig. 14 Average CMR values versus different block and cell sizes.

The block and cell sizes ($m \times m$ cell blocks of $n \times n$ pixel cells) affect the performance of HOPC$_{ncc}$. Fig. 14 shows the average CMR values versus the different block and cell sizes with a half-block overlap, and Table II lists the average CMR values and run times. It can be seen that the average CMR value drops when the cell size increases. Indeed, 3-4 pixel wide cells achieve the best results irrespective of the block size. In addition, 3×3 cell blocks perform best. The valuable spatial information is suppressed if the block becomes too large or too small, which is unfavorable for image matching. In this analysis, 3×3 cell blocks of 3×3 pixel cells achieves the highest CMR value, followed by 3×3 cell blocks of 4×4 pixel cells. However, the difference between their CMR values is only 0.2%, and the choice of 3×3 cell blocks of 4×4 pixel cells has an obvious advantage in computational efficiency compared to 3×3 cell blocks of 3×3 pixel cells. Therefore, 3×3 cell blocks of 4×4 pixel cells are used as the optimum values in these experiments.

Based on the above results, the following parameters are identified to compute HOPC$_{ncc}$: $\beta$=8 orientation bins; 3×3 cell blocks of 4×4 pixel cells; and $\alpha$=1/2 block width overlap. These parameters have been used in the experiments described in the next subsection.

TABLE II
AVERAGE CMR VALUES AND RUN TIMES AT DIFFERENT
BLOCK AND CELL SIZES

| Cell size (pixels) | Block size (cells) | | | | | | | |
|---|---|---|---|---|---|---|---|---|
| | 1×1 | | 2×2 | | 3×3 | | 4×4 | |
| | CMR | Time | CMR | Time | CMR | Time | CMR | Time |
| 3×3 | 84.0% | 31.4s | 89.4% | 54.8s | **91.6%** | **44.6s** | 91.1% | 50.7s |
| 4×4 | 82.4% | 31.4s | 88.6% | 31.3s | **91.4%** | **28.9s** | 90.3% | 29.1s |
| 5×5 | 76.2% | 14.2s | 87.2% | 21.2s | 88.4% | 16.7s | 87.8% | 20.5s |
| 6×6 | 73.4% | 14.2s | 81.4% | 13.4s | 84.8% | 14.2s | 80.8% | 13.4s |
| 7×7 | 64.9% | 8.5s | 77.0% | 10.5s | 79.6% | 9.8s | 73.9% | 10.4s |

### D. Analysis of Similarity Curve

The similarity curve can qualitatively analyze the matching performance of similarity metrics [47]. In general, the similarity curve is maximal when the CPs are located at the correct matching position. A pair of visible and SAR images with high resolution are used in this experiment. A template window (68×68 pixels) is first selected from the visible image. Then, NCC, MI, HOG$_{ncc}$, and HOPC$_{ncc}$ are calculated within a search window (20×20 pixels) of the SAR image.

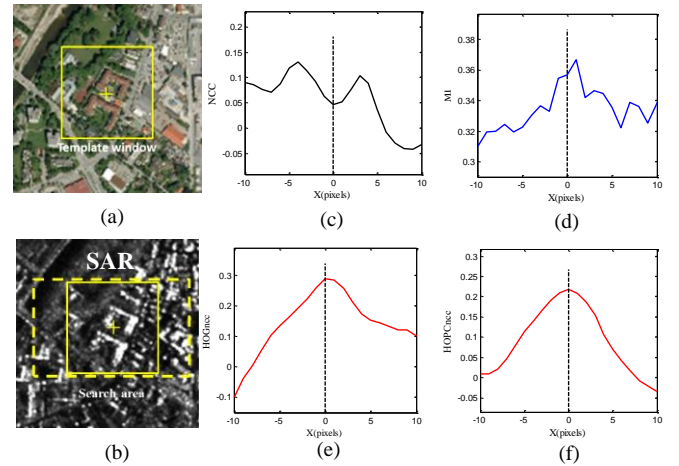

Fig. 15 Similarity curves of NCC, MI, HOG$_{ncc}$, and HOPC$_{ncc}$. (a) Visible image. (b) SAR image. (C) Similarity curve of NCC. (d) Similarity curve of MI. (e) Similarity curve of HOG$_{ncc}$. (f) Similarity curve of HOPC$_{ncc}$.

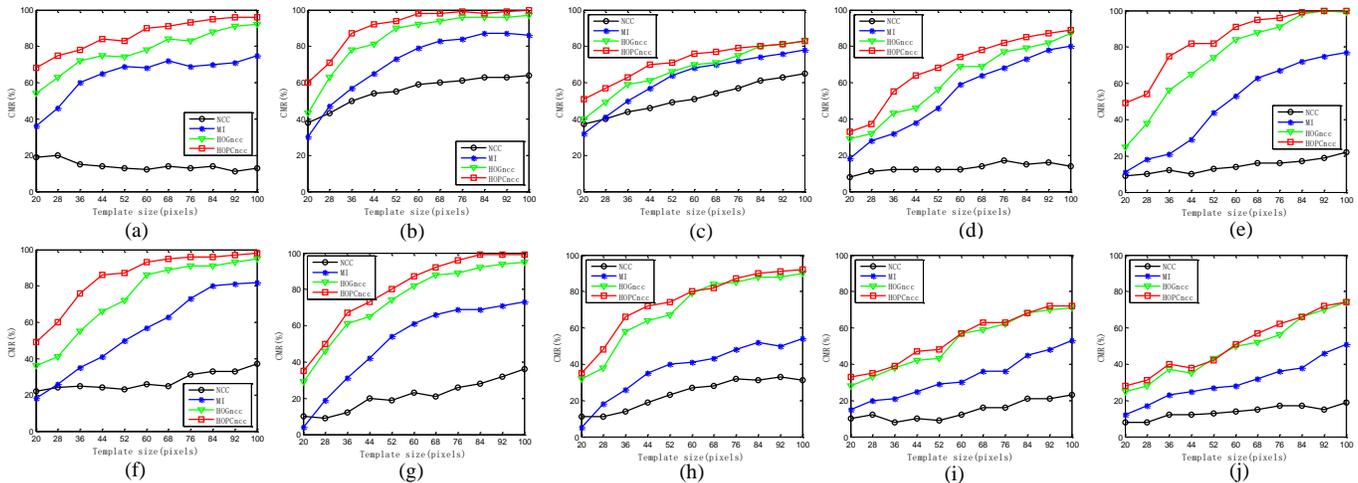

Fig. 18 CMR values versus the template size of NCC, MI, HOG and HOPC$_{ncc}$ for real multimodal images. (a) Visib-Infra 1. (b) Visib-Infra 2. (c) LiDAR-Visib 1. (d) LiDAR-Visib 2. (e) LiDAR-Visib 3. (f) Visib-SAR 1. (g) Visib-SAR 2. (h) Visib-SAR 3. (i) Img-Map 1. (j) Img-Map 2.

Fig. 15 shows the similarity curves of NCC, MI, HOG$_{ncc}$, and HOPC$_{ncc}$. One can clearly see that the significant radiometric differences cause both NCC and MI to fail to detect the CP. Even if HOG$_{ncc}$ achieves the correct CP at the maximum, its curve peak is not very significant. By comparison, HOPC$_{ncc}$ not only detects the correct CP, but also exhibits a smoother similarity curve and more distinguishable peak. This example indicates that HOPC$_{ncc}$ is more robust than the other similarity metrics to the non-linear radiometric differences. A more detailed analysis of the performance of HOPC$_{ncc}$ is provided in the next subsections.

### E. Analysis of Correct Matching Ratio

In this subsection, the performance of NCC, MI, HOG$_{ncc}$, and HOPC$_{ncc}$ is evaluated by using the synthetic and real datasets in terms of CMR. In the matching processing, template windows of different sizes (from 20×20 to 100×100 pixels) are used to detect the CPs for analyzing the sensitivity of these similarity metrics with respect to changes in the template size.

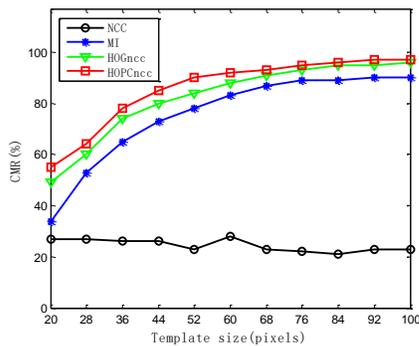

Fig. 16 CMR values versus the template size for the synthetic image pair.

#### Results on Synthetic Datasets

Fig. 16 shows the CMR values versus the template size for the synthetic image pair with non-linear radiometric differences. It can be clearly seen that HOPC$_{ncc}$ performs best in any template size, followed by HOG$_{ncc}$ and MI, whereas NCC achieves the lowest CMR values because it is vulnerable to the piecewise linear intensity mapping used to generate the synthetic images [19, 20]. Moreover, the CMR values of HOPC$_{ncc}$, HOG$_{ncc}$, and MI increase as the template size increases, while NCC does not present a similar regularity. Compared with HOG$_{ncc}$, HOPC$_{ncc}$ exhibits a slight superiority because the simulated radiometric distortion models yield the non-uniform illumination and contrast changes between images (see Fig. 11), and the phase congruency feature (used for HOPC$_{ncc}$) is more robust to these changes compared with gradient information (used for HOG$_{ncc}$). Fig. 17 shows the CPs detected using HOPC$_{ncc}$ with a template size of 100×100 pixels between the synthetic images. In the enlarged sub-images, one can see that the CPs are correctly located in the exact positions.

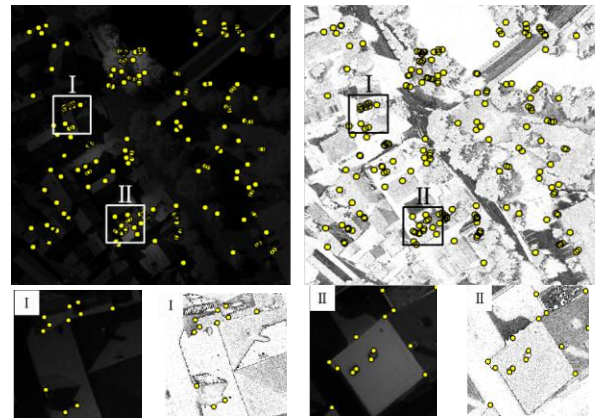

Fig. 17 CPs detected by HOPC$_{ncc}$ with the template size of 100×100 pixels (synthetic images).

#### Results on Real Datasets

To comprehensively evaluate the proposed similarity metric in a real situation, experiments also are performed on different kinds of multimodal remote sensing images (Visib-Infra, LiDAR-Visib, Visib-SAR, and Img-Map). The performance of the similarity metrics for different kinds of image pairs mainly depends on the radiometric distortions between each pair of images. In general, the matching of Visib-SAR and Img-Map is more difficult than that of Visib-Infra due to the presence of more significant radiometric differences and noises.

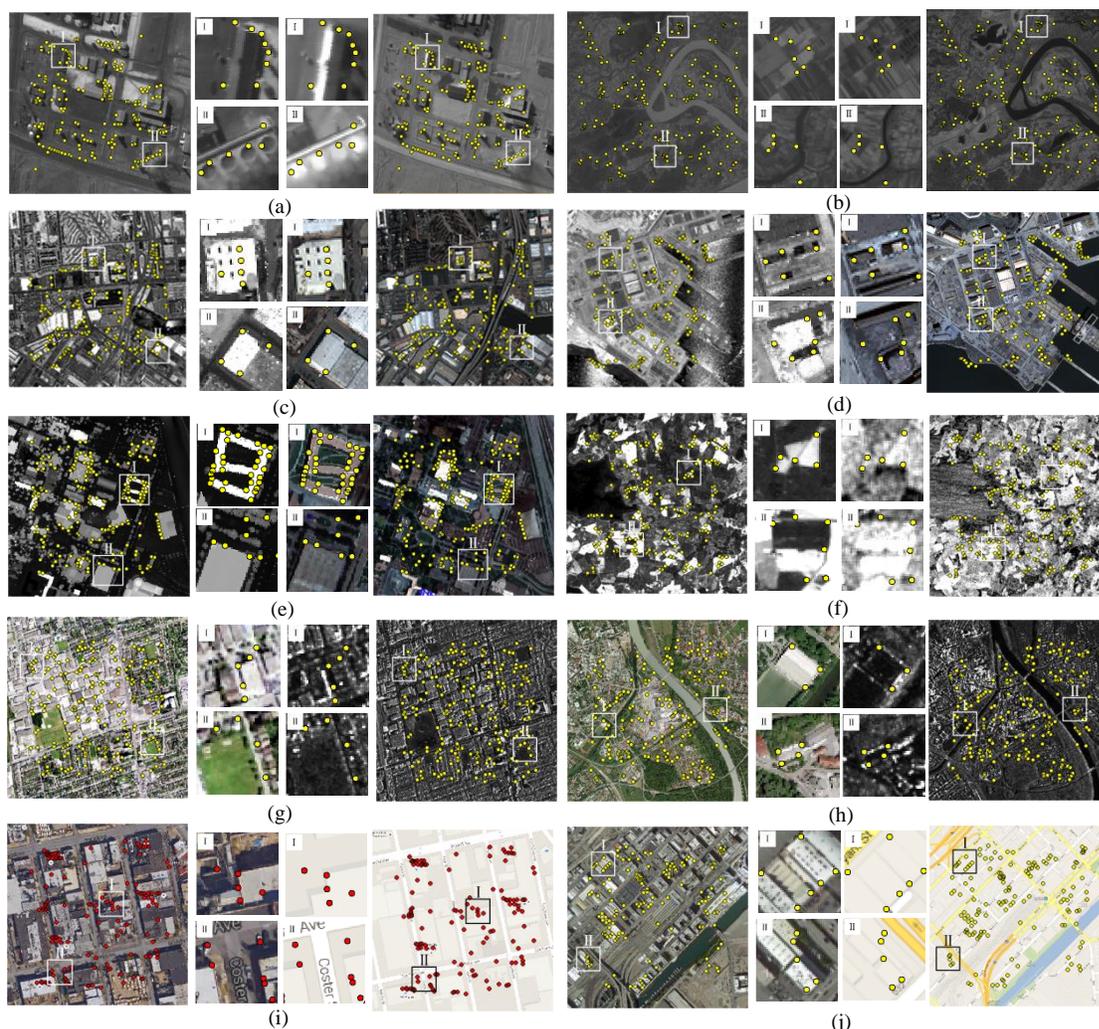

Fig. 19 CPs identified by HOPC_ncc with the template size of 100×100 pixels (real images). (a) Visib-Infra 1. (b) Visib-Infra 2. (c) LiDAR-Visib 1. (d) LiDAR-Visib 2. (e) LiDAR-Visib 3. (f) Visib-SAR 1. (g) Visib-SAR 2. (h) Visib-SAR 3. (i) Img-Map 1. (j) Img-Map 2.

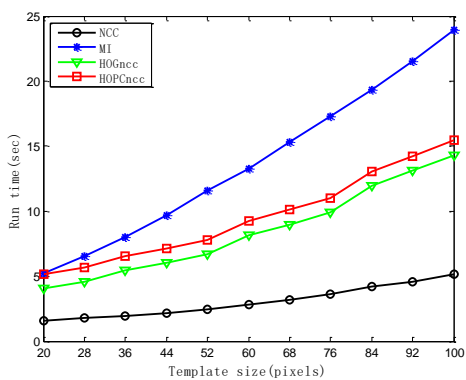

Fig. 2 Run time versus the template size to NCC, MI, HOG_ncc and HOPC_ncc.

Fig. 18 shows the comparative CMR values of the four similarity metrics for the real multimodal images. In almost all the tests, HOPC_ncc outperforms the other similarity metrics for any template size, and HOG_ncc achieves the second highest

CMR values, followed by MI. In contrast, NCC is quite sensitive for multimodal images and achieves the lowest CMR values compared with the other similarity metrics.

Apart from having higher CMR values, the performance of HOPCncc is less affected by template size compared with MI. Taking LiDAR-Visib 3 as an example [see Fig.19(e)], the performance of MI is very sensitive to template size changes, and its CMR value is less than 25% when the template size is small (less than 36×36 pixels). In contrast, HOPC_ncc achieves a CMR value of 75%. The reason for this behavior is that MI computes the joint entropy between images, which is quite sensitive to sample sizes (i.e., template sizes) [20]. In addition, HOPC_ncc performs much better than MI for the high-resolution multimodal images (LiDAR-Visib 3, Visib-SAR 2 and 3, and Img-Map 1 and 2). As shown in Fig. 18 (h), the CMR value of HOPC_ncc reaches 92%, while MI has a CMR value of only 54.5% with a large template size (100×100 pixels). Similar results are shown in Fig. 18 (g), (i) and (j). These results are mainly due to the fact that high-resolution images usually have salient structural features. Thus, HOPC_ncc representing the structural similarity has an obvious superiority to MI.

In the experiments, HOPC_ncc and HOG_ncc have achieved the two highest CMR values, which confirms that the similarity metrics capturing structural properties are more robust to the nonlinear radiometric differences than the other similarity metrics. HOPC$_{ncc}$ exhibits better performance than HOG$_{ncc}$ because HOPC$_{ncc}$ is based on phase congruency, which is more robust to radiometric distortions (illumination and contrast changes) than the gradients used to build HOG$_{ncc}$.

TABLE III
DESCRIPTION OF DATASETS USED IN THE MULTIMODAL REGISTRATION EXPERIMENTS

| Category | No. | Dataset description | | |
|---|---|---|---|---|
| | | Master image | Slave image | Image characteristic |
| Visib-Infra | 1 | Sensor: SPOT 4 band 2<br>GSD: 30m<br>Date: 09/2002<br>Size: 1475×1485 | Sensor: Landsat 5 TM band5<br>GSD: 30m<br>Date: 04/2000<br>Size: 973×988 | Images cover a suburban area located in the south part of Wuhan, China. There is a temporal difference of 29 months between the images |
| Img-Map | 1 | Source: Google Maps<br>GSD: 1m<br>Date: unknown<br>Size: 1337×1369 | Source: Google Maps<br>GSD: 1m<br>Date: unknown<br>Size: 1353×1369 | Images cover an urban area located in Foster City, USA. Their intensity information are largely different |
| SAR-Visib | 1 | Sensor: TerraSAR-X<br>GSD: 30m<br>Date: 03/2008<br>Size: 1138×1251 | Sensor: TM band3<br>GSD: 30m<br>Date: 05/2007<br>Size: 1128×1251 | Images cover a suburban area located in Rugen, Germany. The images have the significant radiometric differences. |
| SAR-Visib | 2 | Sensor: TerraSAR-X<br>GSD: 3m<br>Date: 01/2008<br>Size: 1169×1221 | Source: Google Earth<br>GSD: 3m<br>Date: 03/2009<br>Size: 1006×1123 | Images cover an urban area located in Rosenheim, Germany. The images have significant radiometric differences and local distortions. Moreover, they have a temporal difference of 14 months. |
| LiDAR-Visib | 1 | Sensor: LiDAR height<br>GSD: 2m<br>Date: 10/2010<br>Size: 915×936 | Sensor: WorldView 2<br>GSD: 2m<br>Date: 10/2011<br>Size: 976×992 | Images cover an urban area with high buildings located in San Francisco, USA. The images have significant radiometric differences and local distortions. Moreover, they have a temporal difference of 12 months, and the LiDAR height image is affected by significant noise. |
| LiDAR-Visib | 2 | Sensor: LiDAR intensity<br>GSD: 2m<br>Date: 10/2010<br>Size: 1319×1383 | Sensor: WorldView 2<br>GSD: 2m<br>Date: 10/2011<br>Size: 1195×1223 | Images cover an urban area with high buildings located in San Francisco, USA. The images have significant radiometric differences and local distortions. Moreover, they have a temporal difference of 12 months, and the LiDAR intensity image is affected by significant noise. |

All the above results demonstrate the effectiveness and advantage of the proposed structural similarity metric in the matching performance. The CPs detected by using HOPC$_{ncc}$ on all the real multimodal images are shown in Fig. 19.

### F. Analysis of Computational Efficiency

Computational efficiency is another important indicator for evaluating the matching performance of similarity metrics. Fig. 20 shows the run time taken from NCC, MI, HOG$_{ncc}$, and HOPC$_{ncc}$ versus the template size. HOPC$_{ncc}$ and HOG$_{ncc}$ are both calculated by the proposed fast matching scheme (see Section II-F). The experiments have been performed on an Intel Core i7-4710MQ 2.50GHz PC. One can see that NCC requires the least amount of run time among the similarity metrics due to its lowest computational complexity [20]. Since HOPC$_{ncc}$ and HOG$_{ncc}$ need to extract the structural descriptors and calculate the NCC between such descriptors, they are both more time-consuming than NCC. However, their computational efficiency is better than that of MI because MI calculates the joint histogram for every matched template window pair, which requires extensive computation [19]. The results depicted in Fig. 20 illustrate that HOPC$_{ncc}$ requires more run time than HOG$_{ncc}$ mainly because HOPC$_{ncc}$ is required to extract the phase congruency feature, which is more time-consuming than calculating the gradients used to construct HOG$_{ncc}$.

## V. EXPERIMENTAL RESULTS: MULTIMODAL REGISTRATION

To validate the effectiveness of the proposed registration method based on HOPC$_{ncc}$ (see Section III), manual registration and a registration method based on SIFT are used for comparison. In the proposed method, the block-based Harris operator is set to extract 300 evenly distributed interest points for image registration. For manual registration, 30 CPs are selected evenly over the master and slave images, and the PL transformation model is applied to achieve image registration. In the SIFT-based registration, the feature points are first extracted from both images through the SIFT algorithm, then a one-to-one matching between feature points is performed using the Euclidean distance ratio between the first and the second nearest neighbor. RANdom SAmple Consensus [48] is used to remove the outliers to achieve the final CPs. Finally, the slave image is rectified by the PL transformation model. To assess the registration accuracy, 40-60 check points are selected evenly between the master and registered images by the method described in Section IV-B, and the RMSE of check points is used for accuracy evaluation.

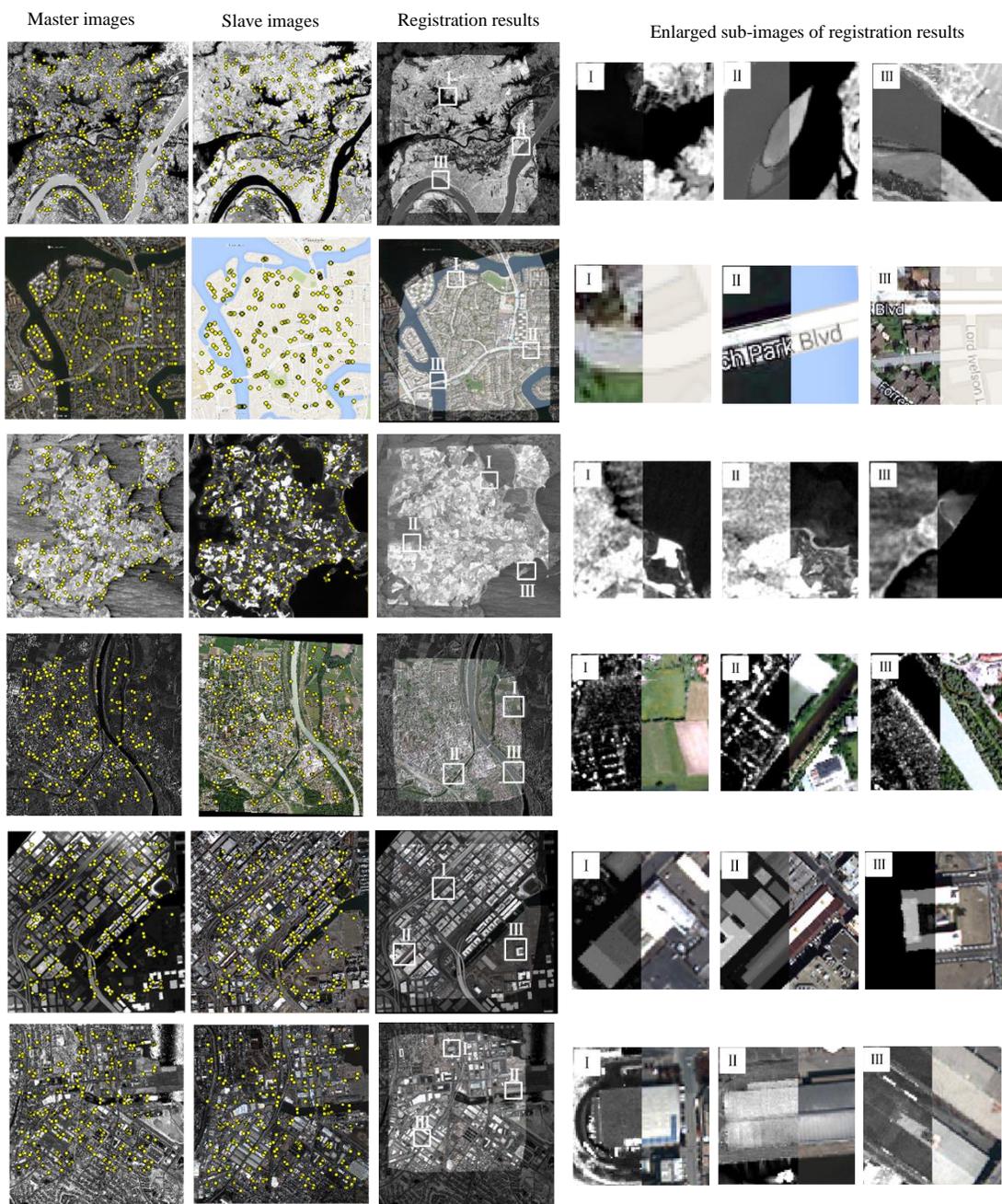

Fig. 21 Registration results for all the test sets. The lines 1, 2,3,4,5, and 6 correspond to Visib-Infra 1, Img-Map 1, SAR-Visib 1, SAR-Visib 2, LiDAR-Visib 1, and LiDAR-Visib 2, respectively.

### A. Description of Datasets

Six sets of multimodal images are used to validate the proposed method. Also for these experiments, the test sets include various kinds of multimodal images such as Visib-Infra, LiDAR-Visib, SAR-Visib, and Img-Map. The master and slave images of each test are captured by different sensors and at different spectral regions, which results in significant non-linear radiometric differences. The description of datasets is given in Table III.

### B. Registration Results

Table IV reports the registration accuracies for the six test sets. The proposed method is successful in registering all the image pairs and achieves the highest registration accuracy. For the SAR-to-Visible and LiDAR-to-Visible image registration (SAR-Visib 1, SAR-Visib 2, LiDAR-Visib 1, and LiDAR-Visib 2), the proposed method outperforms manual registration. One reason for this outcome is that the image pairs of these test sets have significant radiometric differences and the SAR and LiDAR data contain significant noise, which results in a large difference between the intensity details of the two images and makes it difficult to locate the CPs precisely by visual inspection. Another reason is that the proposed method detects many more CPs than manual registration, which is very beneficial to the PL transformation model for fitting complex deformations between images [43]. In addition, the SIFT-based registration fails in most of the tests except Visib-Infra 1 because the SIFT algorithm is not able to extract the highly-repeatable common features present in multimodal images due to their significant radiometric differences [49].

On the other hand, one can see that the proposed method achieves different registration accuracies for different test sets because of the differences in the image characteristics. In general, the test sets having images with lower resolutions achieve relatively higher registration accuracy than those with higher resolutions. For example, Visib-Infra 1 and SAR-Visib 1 achieve a sub-pixel registration accuracy, whereas the other test sets have a RMSE larger than 1 pixel. This is mainly attributed to the fact that the test sets which include the images with low resolutions cover flat areas and there is almost no complex geometric deformation between the images. The higher resolution images covering urban areas, such as the image pairs of SAR-Visib 2, LiDAR-Visib 1 and 2, have significant local distortions caused by relief displacement of buildings. This is an intrinsic problem for high-resolution registration, which cannot be resolved by an image-to-image registration until a true orthorectification is applied [50]. Fig. 21 shows the registration results of all the test sets. From the enlarged sub-images, one can see that the registrations are satisfactory and accurate for all the test sets. The above results demonstrate the effectiveness of the proposed method for registering multimodal remote sensing images.

TABLE IV
REGISTRATION RESULTS FOR ALL THE CONSIDERED TEST SETS

| Category | No. | Method | Matched CPs | CPs with error elimination | RMSE (pixels) |
|---|---|---|---|---|---|
| Visib-Infra | 1 | Proposed | 278 | 278 | 0.668 |
| | | Manual | 30 | 30 | 0.937 |
| | | SIFT | 126 | 51 | 1.345 |
| Img-Map | 1 | Proposed | 256 | 246 | 1.056 |
| | | Manual | 30 | 30 | 2.084 |
| | | SIFT | 115 | 0 | Failed |
| SAR-Visib | 1 | Proposed | 289 | 279 | 0.765 |
| | | Manual | 30 | 30 | 1.746 |
| | | SIFT | 108 | 0 | Failed |
| | 2 | Proposed | 216 | 215 | 1.206 |
| | | Manual | 30 | 30 | 2.290 |
| | | SIFT | 81 | 0 | Failed |
| LiDAR-Visib | 1 | Proposed | 289 | 285 | 1.256 |
| | | Manual | 30 | 30 | 2.389 |
| | | SIFT | 121 | 0 | Failed |
| | 2 | Proposed | 225 | 216 | 1.314 |
| | | Manual | 30 | 30 | 2.118 |
| | | SIFT | 276 | 0 | Failed |

## VI. CONCLUSION

This paper has presented a novel similarity metric named HOPC$_{ncc}$ for multimodal remote sensing image registration. This metric addresses the issues related to the significant non-linear radiometric differences usually present in images acquired by different sensors. First, the phase congruency model is extended to build its orientation representation. Then, the amplitude and orientation of phase congruency are used to construct HOPC$_{ncc}$, and a fast template matching scheme is designed for this metric to detect CPs. HOPC$_{ncc}$ aims to capture the structural similarity between images and has been evaluated against various kinds of multimodal datasets, including Visible-to-Infrared, LiDAR-to-Visible, Visible-to-SAR, and Image-to-Map. The experimental results demonstrate clearly that HOPC$_{ncc}$ outperforms the two popular similarity metrics including NCC and MI, especially for image pairs that contained rich structural features, such as the high-resolution visible and SAR images (Visib-SAR 2 and Visib-SAR 3 in Table I), and the LiDAR height and visible images (LiDAR-Visib 3 in Table I). Moreover, when HOPC$_{ncc}$ is implemented with the proposed fast matching scheme, less computation time is required compared to MI. A robust registration method based on HOPC$_{ncc}$ for multimodal images is introduced that uses various techniques including the block-based Harris operator, HOPC$_{ncc}$, bidirectional matching, and PL transformation. The experimental results using six different pairs of multimodal images confirms that the new method can detect a large number of evenly distributed CPs between the images and its registration accuracy is better than the manual and SIFT-based registration methods.

Since HOPC$_{ncc}$ uses the framework of HOG to build the descriptor, the HOG descriptor is also integrated as a similarity metric (named HOG$_{ncc}$) for image registration. The experimental results show that both HOPC$_{ncc}$ and HOG$_{ncc}$ perform better than NCC and MI, which demonstrates that the framework of HOG is effective for building a structural descriptor for multimodal registration. When compared with HOG$_{ncc}$, HOPC$_{ncc}$ improves the matching performance by using phase congruency instead of gradient information to build the descriptor. In future efforts, we will attempt to integrate other features (e.g., wavelets and self-similarity [51,52]) into the framework of HOG for multimodal remote sensing image registration.

Although our experiments show that HOPC$_{ncc}$ is robust to non-linear radiometric differences, some improvements to HOPC$_{ncc}$ should be considered. One limitation of HOPC$_{ncc}$ is that it is not invariant for scale and rotation changes, which could be critical in cases where significant changes of scale and rotation are present between images. In practice, these deformations between images need to be eliminated using the direct georeferencing technique based on the navigation instruments aboard satellites. A Fourier analysis method for rotation-invariant local descriptor [53] may also address this issue to some degree. Although HOPC$_{ncc}$ is applied through a fast matching scheme, it is still more time-consuming compared with NCC since HOPC$_{ncc}$ requires the calculation of a high-dimensional structural feature descriptor to be calculated. In future work, this issue could be resolved by reducing the dimensions of the descriptor using a dimension-reduction technique, such as principal component analysis. In addition, it is worth noting that the performance of HOPC$_{ncc}$ may degrade if the images of interest include less structure or shape information because HOPC$_{ncc}$ depends on the structural properties of images. In this case, an image enhancement approach could be applied to enhance the contour or edge features, which may be helpful to image registration.

## ACKNOWLEDGMENT

The authors would like to thank the anonymous reviewers for their helpful comments and good suggestions.

## REFERENCES

[1] B. Zitova and J. Flusser, "Image registration methods: a survey," *Image Vis. Comput.,* vol. 21, no. 11, pp. 977-1000, Oct. 2003.

[2] H. Goncalves, L. Corte-Real, and J. A. Goncalves, "Automatic image registration through image segmentation and SIFT," *IEEE Trans. Geosci. Remote Sens.,* vol. 49, no. 7, pp. 2589-2600, Mar. 2011.

[3] H. Goncalves, J. A. Goncalves, L. Corte-Real, and A. C. Teodoro, "CHAIR: Automatic image registration based on correlation and Hough transform," *Int. J. Remote Sens.,* vol. 33, no. 24, pp. 7936-7968, Jul. 2012.

[4] P. Bunting, F. Labrosse, and R. Lucas, "A multi-resolution area-based technique for automatic multi-modal image registration," *Image Vis. Comput.,* vol. 28, no. 8, pp. 1203-1219, Aug. 2010.

[5] L. Yu, D. R. Zhang, and E. J. Holden, "A fast and fully automatic registration approach based on point features for multi-source remote-sensing images," *Comput. Geosci.,* vol. 34, no. 7, pp. 838-848, Jul. 2008.

[6] H. G. Sui, C. Xu, J. Y. Liu, and F. Hua, "Automatic optical-to-SAR image registration by iterative line extraction and voronoi integrated spectral point matching," *IEEE Trans. Geosci. Remote Sens.,* vol. 53, no. 11, pp. 6058-6072, Nov. 2015.

[7] H. Gonçalves, J. Gonçalves, and L. Corte-Real, "HAIRIS: A method for automatic image registration through histogram-based image segmentation," *IEEE Trans. Image Process.,* vol. 20, no. 3, pp. 776-789, Sep. 2011.

[8] D. G. Lowe, "Distinctive image features from scale-invariant keypoints," *Int. J. Comput. Vis.,* vol. 60, no. 2, pp. 91-110, Nov. 2004.

[9] K. Mikolajczyk and C. Schmid, "A performance evaluation of local descriptors," *IEEE Trans. Pattern Anal. Mach. Intell.,* vol. 27, no. 10, pp. 1615-1630, Oct. 2005.

[10] J. L. Ma, J. C. W. Chan, and F. Canters, "Fully automatic subpixel image registration of multiangle CHRIS/Proba data," *IEEE Trans. Geosci. Remote Sens.,* vol. 48, no. 7, pp. 2829-2839, Jul. 2010.

[11] A. Sedaghat, M. Mokhtarzade, and H. Ebadi, "Uniform robust scale-invariant feature matching for optical remote sensing images," *IEEE Trans. Geosci. Remote Sens.,* vol. 49, no. 11, pp. 4516-4527, Nov. 2011.

[12] A. Sedaghat, and H. Ebadi, "Remote sensing image matching based on adaptive binning SIFT descriptor," *IEEE Trans. Geosci. Remote Sens.,* vol. 53, no. 10, pp. 5283-5293, Oct. 2015.

[13] Y. Ye, and J. Shan, "A local descriptor-based registration method for multispectral remote sensing images with non-linear intensity differences," *ISPRS J. Photogramm. Remote Sens.,* vol. 90, pp. 83-95, Apr. 2014.

[14] H. Bay, A. Ess, T. Tuytelaars, and L. Van Gool, "Speeded-up robust features (SURF)," *Comput. Vision Image Understanding,* vol. 110, no. 3, pp. 346-359, Jun. 2008.

[15] E. Rublee, V. Rabaud, K. Konolige, and G. Bradski, "ORB: An efficient alternative to SIFT or SURF," in *Proc. IEEE Int. Conf. Comput. Vis.,* 2011, pp. 2564–2571.

[16] A. Alahi, R. Ortiz, and P. Vandergheynst, "FREAK: Fast retina keypoint," in *Proc. IEEE Comput. Vis. Pattern Recognit.,* 2012, pp. 510-517.

[17] A. Kelman, M. Sofka, and C. V. Stewart, "Keypoint descriptors for matching across multiple image modalities and non-linear intensity variations," in *Proc. IEEE Comput. Vis. Pattern Recognit.,* 2007, pp. 3257-3263.

[18] J. Inglada and A. Giros, "On the possibility of automatic multisensor image registration," *IEEE Trans. Geosci. Remote Sens.,* vol. 42, no. 10, pp. 2104-2120, Oct. 2004.

[19] Y. Hel-Or, H. Hel-Or, and E. David, "Fast template matching in non-linear tone-mapped images," in *Proc. IEEE Int. Conf. Comput. Vis.,* 2011, pp. 1355-1362.

[20] Y. Hel-Or, H. Hel-Or, and E. David, "Matching by tone mapping: Photometric invariant template matching," *IEEE Trans. Pattern Anal. Mach. Intell.,* vol. 36, no. 2, pp. 317-330, Feb. 2014.

[21] A. A. Cole-Rhodes, K. L. Johnson, J. LeMoigne, and I. Zavorin, "Multiresolution registration of remote sensing imagery by optimization of mutual information using a stochastic gradient," *IEEE Trans. Image Process.,* vol. 12, no. 12, pp. 1495-1511, Dec. 2003.

[22] H. M. Chen, M. K. Arora, and P. K. Varshney, "Mutual information-based image registration for remote sensing data," *Int. J. Remote Sens.,* vol. 24, no. 18, pp. 3701-3706, Sep. 2003.

[23] D. Brunner, G. Lemoine, and L. Bruzzone, "Earthquake damage assessment of buildings using VHR optical and SAR imagery," *IEEE Trans. Geosci. Remote Sens.,* vol. 48, no. 5, pp. 2403-2420, May 2010.

[24] M. Ravanbakhsh, and C. S. Fraser, "A comparative study of DEM registration approaches," *J. Spat. Sci.,* vol. 58, no. 1, pp. 79-89, Apr. 2013.

[25] J. M. Murphy, J. Le Moigne, and D. J. Harding, "Automatic image registration of multimodal remotely sensed data with global shearlet features," *IEEE Trans. Geosci. Remote Sens.,* vol. 54, no. 3, pp. 1685-1704, Mar. 2016.

[26] I. Zavorin and J. Le Moigne, "Use of multiresolution wavelet feature pyramids for automatic registration of multisensor imagery," *IEEE Trans. Image Process.,* vol. 14, no. 6, pp. 770-782, Jun. 2005.

[27] M. P. Heinrich, M. Jenkinson, M. Bhushan, T. Matin, F. V. Gleeson, J. M. Brady, and J. A. Schnabel, "Non-local shape descriptor: a new similarity metric for deformable multi-modal registration," in *Medical Image Computing and Computer-Assisted Intervention,* 2011, pp. 541-548.

[28] M. P. Heinrich, M. Jenkinson, M. Bhushan, T. Matin, F. V. Gleeson, S. M. Brady, and J. A. Schnabel, "MIND: Modality independent neighbourhood descriptor for multi-modal deformable registration," *Med. Image Anal.,* vol. 16, no. 7, pp. 1423-1435, Oct. 2012.

[29] H. Rivaz, Z. Karimaghaloo, and D. L. Collins, "Self-similarity weighted mutual information: A new nonrigid image registration metric," *Med. Image Anal.,* vol. 18, no. 2, pp. 343-358, Feb. 2014.

[30] Z. Li, D. Mahapatra, J. A. W. Tielbeek, J. Stoker, L. J. van Vliet, and F. M. Vos, "Image registration based on autocorrelation of local structure," *IEEE Trans. Med. Imaging,* vol. 35, no. 1, pp. 63-75, Jan. 2016.

[31] P. Kovesi, "Image features from phase congruency," *Videre: Journal of Computer Vision Research,* vol. 1, no. 3, pp. 1-26, 1999.

[32] Y. Ye and L. Shen, "HOPC: A novel similarity metric based on geometric structural properties for multi-modal remote sensing image matching," *ISPRS Ann. Photogramm. Remote Sens. Spatial Inf. Sci.,* pp. 9-16, Jun. 2016.

[33] N. Dalal and B. Triggs, "Histograms of oriented gradients for human detection." in *Proc. IEEE Comput. Vis. Pattern Recognit.,* 2005, pp. 886-893.

[34] J. Canny, "A computational approach to edge-detection," *IEEE Trans. Pattern Anal. Mach. Intell.,* vol. 8, no. 6, pp. 679-698, Nov. 1986.

[35] A. V. Oppenheim and J. S. Lim, "The importance of phase in signals," *Proceedings of the IEEE,* vol. 69, no. 5, pp. 529-541, May 1981.

[36] M. C. Morrone and R. A. Owens, "Feature detection from local energy," *Pattern Recognition Letters,* vol. 6, no. 5, pp. 303-313, Dec. 1987.

[37] P. Moreno, A. Bernardino, and J. Santos-Victor, "Improving the SIFT descriptor with smooth derivative filters," *Pattern Recognition Letters,* vol. 30, no. 1, pp. 18-26, Jan. 2009.

[38] P. F. Felzenszwalb, R. B. Girshick, D. McAllester, and D. Ramanan, "Object detection with discriminatively trained part-based models," *IEEE Trans. Pattern Anal. Mach. Intell.,* vol. 32, no. 9, pp. 1627-1645, 2010.

[39] H. Harzallah, F. Jurie, and C. Schmid, "Combining efficient object localization and image classification," in *Proc. IEEE Int. Conf. Comput. Vis.,* 2009, pp. 237-244.

[40] A. Shrivastava, T. Malisiewicz, A. Gupta, and A. A. Efros, "Data-driven visual similarity for cross-domain image matching," *ACM Trans. Graph.*, vol. 30, no. 6, Dec. 2011.

[41] L. Di Stefano, M. Marchionni, and S. Mattoccia, "A fast area-based stereo matching algorithm," *Image Vis. Comput.*, vol. 22, no. 12, pp. 983-1005, Oct. 2004.

[42] A. Wong and D. A. Clausi, "ARRSI: Automatic registration of remote-sensing images," *IEEE Trans. Geosci. Remote Sens.*, vol. 45, no. 5, pp. 1483-1493, May 2007.

[43] A. Goshtasby, "Piecewise linear mapping functions for image registration," *Pattern Recognit.*, vol. 19, no. 6, pp. 459-466, 1986.

[44] V. J. D. Tsai, "Delaunay triangulations in TIN creation: An overview and a linear-time algorithm," *Int. J. Geogr. Inf. Sci.*, vol. 7, no. 6, pp. 501-524, 1993.

[45] Y. Han, J. Choi, Y. Byun, and Y. Kim, "Parameter optimization for the extraction of matching points between high-resolution multisensor images in urban areas," *IEEE Trans. Geosci. Remote Sens.*, vol. 52, no. 9, pp. 5612-5621, Sep. 2014.

[46] A. Myronenko and X. B. Song, "Intensity-based image registration by minimizing residual complexity," *IEEE Transactions on Medical Imaging*, vol. 29, no. 11, pp. 1882-1891, Nov. 2010.

[47] J. Pluim, J. Maintz, and M. Viergever, "Image registration by maximization of combined mutual information and gradient information." in *Medical Image Computing and Computer-Assisted Intervention*, 2000, pp. 103-129.

[48] M. A. Fischler and R. C. Bolles, "Random sample consensus: A paradigm for model fitting with applications to image analysis and automated cartography," *Communications of the ACM*, vol. 24, no. 6, pp. 381-395, 1981.

[49] C. L. Tsai, C. Y. Li, G. Yang, and K. S. Lin, "The edge-driven dual-bootstrap iterative closest point algorithm for registration of multimodal fluorescein angiogram sequence," *IEEE Trans. Med. Imaging*, vol. 29, no. 3, pp. 636-49, Mar. 2010.

[50] G. Hong and Y. Zhang, "Wavelet-based image registration technique for high-resolution remote sensing images," *Comput. Geosci.*, vol. 34, no. 12, pp. 1708-1720, Dec. 2008.

[51] E. Shechtman and M. Irani, "Matching local self-similarities across images and videos," in *Proc. IEEE Comput. Vis. Pattern Recognit.*, 2007, pp. 1-8.

[52] Y. Ye, L. Shen, M. Hao, J. Cheng, and Z. Xu, "Robust Optical-to-SAR Image Matching Based on Shape Properties," *IEEE Geosci. Remote Sens. Lett.*, 2017 (in press).

[53] K. Liu, H. Skibbe, T. Schmidt, T. Blein, K. Palme, T. Brox, and O. Ronneberger, "Rotation-invariant HOG descriptors using Fourier analysis in polar and spherical coordinates," *Int. J. Comput. Vis.*, vol. 106, no. 3, pp. 342-364, Feb. 2014.

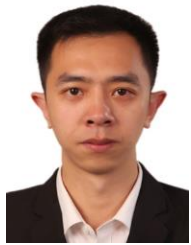

**Yuanxin Ye** (M'17) received the B.S. degree in Remote Sensing Science and Technology from Southwest Jiaotong University, Chengdu, China, in 2008, and the Ph.D. degree in Photogrammetry and Remote Sensing from Wuhan University, Wuhan, China, in 2013. Since Sep. 2013, he has been an Assistant Professor with the Faculty of Geosciences and Environmental Engineering, Southwest Jiaotong University, Chengdu, China. He is currently a postdoctoral fellow in the Remote Sensing Laboratory in the Department of Information Engineering and Computer Science, University of Trento. He achieved "The ISPRS Prizes for Best Papers by Young Authors" of 23th International Society for Photogrammetry and Remote Sensing Congress (Prague, July 2016). His research interests include remote sensing image processing, image registration, feature extraction, and change detection.

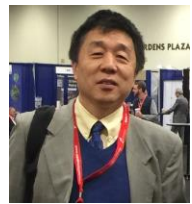

**Jie Shan** (SM'14) received his Ph.D. degree in photogrammetry and remote sensing from Wuhan University, China. He has held faculty positions at universities in China and Sweden, and has been a Research Fellow in Germany. He is with the Lyles School of Civil Engineering, Purdue University, West Lafayette, IN, USA. His areas of interests include sensor geometry, pattern recognition from images and light detection and ranging (LiDAR) data, object extraction and reconstruction, urban remote sensing, and automated mapping. He is an Associate Editor for the IEEE Transactions on Geoscience and Remote Sensing. He is a recipient of multiple academic awards, including the Talbert Abrams Grand Award and the Environmental Systems Research Institute Award for Best Scientific Paper in Geographic Information Systems (First Place).

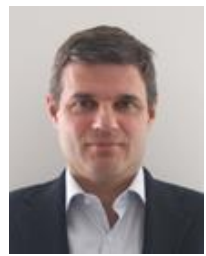

**Lorenzo Bruzzone** (S'95 - M'98 - SM'03 - F'10) received the Laurea (M.S.) degree in electronic engineering (summa cum laude) and the Ph.D. degree in telecommunications from the University of Genoa, Italy, in 1993 and 1998, respectively. He is currently a Full Professor of telecommunications at the University of Trento, Italy, where he teaches remote sensing, radar, pattern recognition, and electrical communications. Dr. Bruzzone is the founder and the director of the Remote Sensing Laboratory in the Department of Information Engineering and Computer Science, University of Trento. His current research interests are in the areas of remote sensing, radar and SAR, signal processing, and pattern recognition. He promotes and supervises research on these topics within the frameworks of many national and international projects. Among the others, he is the Principal Investigator of the Radar for icy Moon exploration (RIME) instrument in the framework of the JUpiter ICy moons Explorer (JUICE) mission of the European Space Agency. He is the author (or coauthor) of 186 scientific publications in referred international journals (134 in IEEE journals), more than 260 papers in conference proceedings, and 20 book chapters. He is editor/co-editor of 16 books/conference proceedings and 1 scientific book. His papers are highly cited, as proven form the total number of citations (more than 16800) and the value of the h-index (66) (source: Google Scholar). He was invited as keynote speaker in 30 international conferences and workshops. Since 2009 he is a member of the Administrative Committee of the IEEE Geoscience and Remote Sensing Society. Dr. Bruzzone ranked first place in the Student Prize Paper Competition of the 1998 IEEE International Geoscience and Remote Sensing Symposium (Seattle, July 1998). Since that time he was recipient of many international and national honors and awards. Dr. Bruzzone was a Guest Co-Editor of different Special Issues of international journals. He is the co-founder of the IEEE International Workshop on the Analysis of Multi-Temporal Remote-Sensing Images (MultiTemp) series and is currently a member of the Permanent Steering Committee of this series of workshops. Since 2003 he has been the Chair of the SPIE Conference on Image and Signal Processing for Remote Sensing. Since 2013 he has been the founder Editor-in-Chief of the IEEE Geoscience and Remote Sensing Magazine. Currently he is an Associate Editor for the IEEE Transactions on Geoscience and Remote Sensing and the Journal of Applied Remote Sensing. Since 2012 he has been appointed Distinguished Speaker of the IEEE Geoscience and Remote Sensing Society.

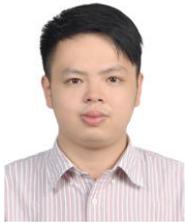

**Li Shen** received the B.S. degree in Photogrammetry and Remote Sensing from Wuhan University, Wuhan, China, in 2008, and the Ph.D. degree from the College of Resources Science and Technology, Beijing Normal University, Beijing, China, in 2013. He is currently an Assistant Professor with the Faculty of Geosciences and Environmental Engineering, Southwest Jiaotong University, Chengdu, China. His research interests include remote sensing image processing, pattern recognition, and remote sensing applications in natural disaster reduction and detection of potential hazards along the high-speed railway line.